\documentclass[letterpaper, 10 pt, conference]{ieeeconf}  

\IEEEoverridecommandlockouts                              

\usepackage{algorithm}
\usepackage{algorithmic}
\usepackage[bookmarks=true]{hyperref}
\usepackage{hyperref}
\usepackage{amsmath} 
\usepackage{amssymb}  
\usepackage{booktabs}
\usepackage{graphicx}
\usepackage{multicol}
\usepackage{multirow}
\usepackage[T1]{fontenc}
\usepackage{colortbl}
\usepackage{hyperref}
\usepackage[doi=false,isbn=false,url=false,eprint=false]{biblatex}

\usepackage{color}
\usepackage[usenames,dvipsnames,table]{xcolor}

\hypersetup{
    colorlinks=true,
    linkcolor=MidnightBlue,
    filecolor=magenta,      
    urlcolor=MidnightBlue,
    citecolor=MidnightBlue,
} 
\usepackage[font=small,labelfont=bf]{caption}

\usepackage{subcaption}

\title{\LARGE \bf
Knowledge-Driven Imitation Learning:\\ Enabling Generalization Across Diverse Conditions
}

\author{Zhuochen Miao$^{*1}$, Jun Lv$^{*1,3}$, Hongjie Fang$^{1}$, Yang Jin$^{1}$, Cewu Lu$^{1,2,3}$
\\
\texttt{\url{knowledge-driven.github.io}}
\thanks{*Equal Contribution.}
\thanks{$^{1}$Shanghai Jiao Tong University.}%
\thanks{$^{2}$Shanghai Innovation Institution.}
\thanks{$^{3}$Shanghai Noematrix Intelligence Technology Ltd.}
}

\begin{document}

\maketitle
\thispagestyle{empty}
\pagestyle{empty}

\begin{abstract}

Imitation learning has emerged as a powerful paradigm in robot manipulation, yet its generalization capability remains constrained by object-specific dependencies in limited expert demonstrations. To address this challenge, we propose knowledge-driven imitation learning, a framework that leverages external structural semantic knowledge to abstract object representations within the same category. We introduce a novel semantic keypoint graph as a knowledge template and develop a coarse-to-fine template-matching algorithm that optimizes both structural consistency and semantic similarity. Evaluated on three real-world robotic manipulation tasks, our method achieves superior performance, surpassing image-based diffusion policies with only one-quarter of the expert demonstrations. Extensive experiments further demonstrate its robustness across novel objects, backgrounds, and lighting conditions. This work pioneers a knowledge-driven approach to data-efficient robotic learning in real-world settings. Code and more materials are available on \url{knowledge-driven.github.io}.

\end{abstract}

\section{Introduction}

In recent years, the robot learning community has witnessed significant progress in imitation learning~\cite{brohan2023rt, chi2024diffusionpolicy, mandlekar2021matters, wang2024rise, aloha_act, rt2}. By mimicking behaviors from human demonstrations, robots have demonstrated remarkable performance across a wide range of manipulation tasks. Despite these advancements, a key challenge remains: imitation policies often struggle to generalize beyond their training environments, leading to performance degradation under novel conditions, such as when encountering a new object. To address this, a common approach is to improve the generalization performance of the policies by scaling up training demonstrations~\cite{black2024pi_0, openvla, lin2024data, rtx, octo, hpt}. While effective, this strategy requires extensive data collection and substantial computational resources, making it both costly and inefficient.

In contrast, for humans, generalizing the learned skills to different conditions is a natural and effortless process. This ability arises because humans do not memorize actions specific to a single object during the learning process. Instead, they abstract objects into representations that capture object-centric knowledge and semantic information essential for manipulation, while ignoring irrelevant details. These representations allow humans to understand and internalize the underlying logic of the task, thereby mastering the manipulation skills at a fundamental level. When encountering a novel object, humans can establish a connection between the new object and the previously learned abstract representations. Once this connection is made, the learned manipulation skills can be easily transferred to new objects and conditions.

\begin{figure}[t]
    \centering
    \includegraphics[width=1.0\linewidth]{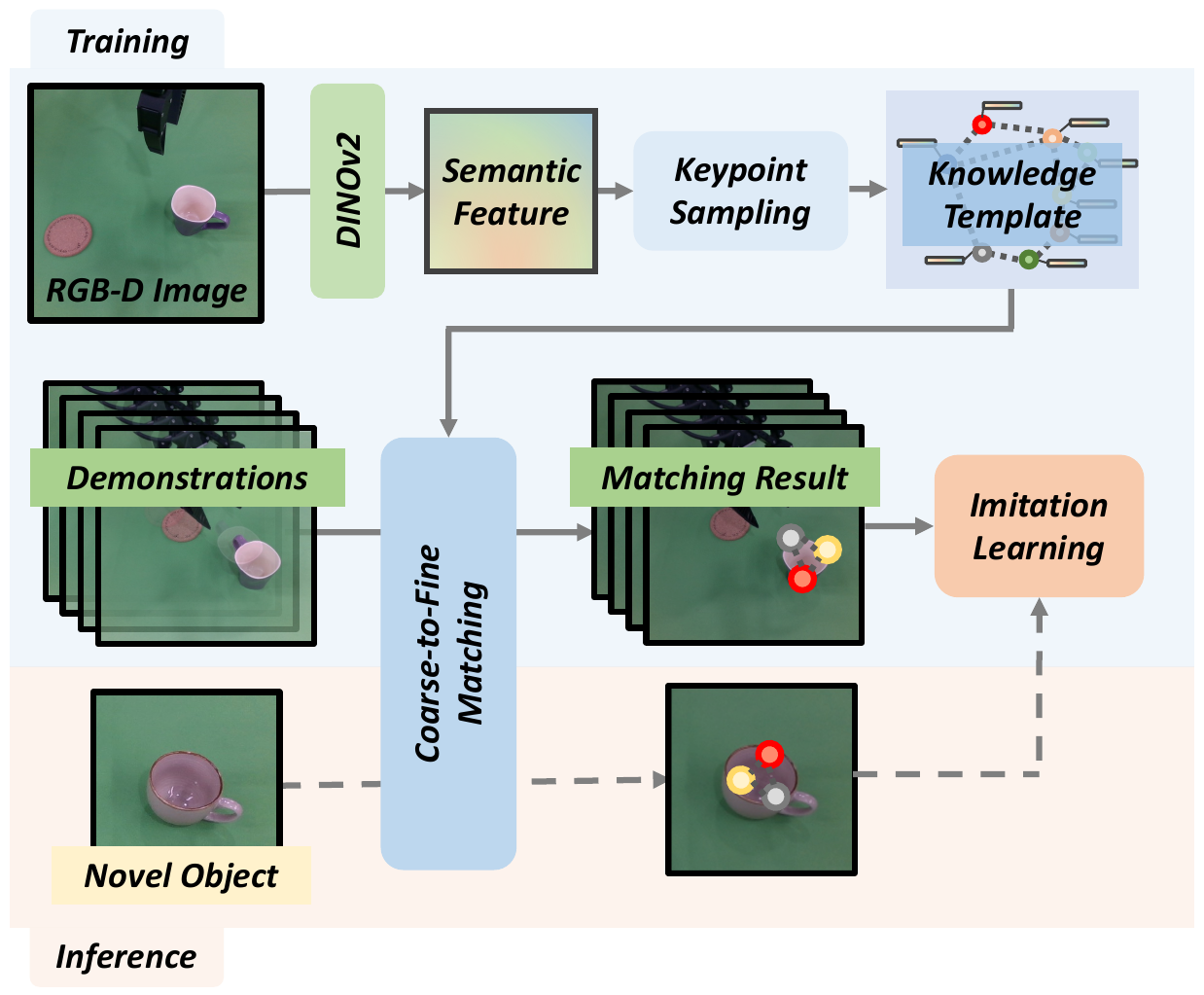}
    \caption{\textbf{Overview of our Knowledge-Driven Imitation Learning Method.} Given an RGB-D image as input, the system generates a knowledge template for a specific object. This template is then matched to the demonstration, and the policy is learned based on the matching results. When encountering novel objects, the learned policy can be transferred, enabling generalization to new scenarios.}
    \label{fig:pipeline}
    \vspace{-5mm}
\end{figure}

Building upon this cognitive insight, we formulate two technical requisites: (1) the development of \textit{structured object-centric knowledge templates} to promote imitation learning efficiency, and (2) the creation of \textit{robust template matching algorithms} capable of generalizing across object instances.

Inspired by these requirements, we propose a knowledge-driven imitation learning system that enhances policy generalization across diverse conditions. The system leverages visual foundation models~\cite{oquab2023dinov2} to abstract task-relevant objects into a semantic knowledge template, represented as a keypoint graph. This graph explicitly encodes object structures to improve the robustness of the matching algorithm while implicitly embedding semantic information within each keypoint. Based on this template, we develop a coarse-to-fine semantic knowledge matching algorithm that abstracts objects into high-level representations, on which the manipulation policy is trained. This approach offers two key advantages: (1) the structured, low-dimensional observation space reduces learning complexity compared to raw sensor inputs, lowering data requirements; and (2) the matching algorithm enables reliable template-object correspondence estimation, supporting skill transfer to novel conditions, such as unseen objects, through consistent knowledge representation alignment.

We validated the effectiveness of our method through real-world experiments on three manipulation tasks. Compared to existing approaches, our method not only achieves higher task success rates across all three tasks but also demonstrates robust performance under variations in objects, scenes, lighting conditions, \textit{etc}. Moreover, our knowledge-driven imitation learning approach significantly improves sample efficiency, surpassing image-based Diffusion Policy~\cite {chi2024diffusionpolicy} with only about 1/4 of the data. Additionally, we quantitatively demonstrate the effectiveness and robustness of our proposed knowledge template and semantic knowledge matching algorithm.

\section{Related Works}

\subsection{Imitation Learning for Robotic Manipulation}

Imitation learning allows robots to acquire manipulation skills by mimicking expert demonstrations. A fundamental approach is behavior cloning (BC)~\cite{bc}, which frames this process as a supervised learning problem, where the agent learns to map observations to actions directly from expert data. Recent advancements~\cite{chi2024diffusionpolicy, florence2022implicit, mandlekar2021matters, shafiullah2022behavior, wang2024rise, aloha_act, fang2025airexo} have further demonstrated the effectiveness of BC in learning from expert demonstrations across various manipulation tasks.

Imitation policies for robotic manipulation typically rely on visual observations, such as 2D images~\cite{chi2024diffusionpolicy, openvla, rdt1b, octo, xia2024cage, aloha_act} or 3D point clouds~\cite{act3d, shridhar2022perceiveractor, dexcap, wang2024rise, dp3}, to generate robot actions. Effective generalization requires the visual encoder to extract robust and transferable representations from the given demonstrations. One approach to enhancing generalization involves scaling up demonstrations~\cite{black2024pi_0, openvla, lin2024data, rtx, octo, hpt} by pre-training on large-scale robotic datasets~\cite{rh20t, droid, rtx, bridge-v2}. In this work, we take a different approach by integrating structured knowledge into the imitation learning pipeline, introducing inductive biases to improve generalization and sample efficiency of robotic manipulation policies.

\subsection{Knowledge-Driven Imitation Learning}

Defining and embedding knowledge into the learning process is crucial for knowledge-driven imitation learning. Several approaches~\cite{spawnnet, soft, xia2024cage, same} have  incorporated knowledge into policies by leveraging visual foundation models~\cite{kirillov2023segany, oquab2023dinov2} as visual encoders to generate generalizable scene-level representations. These representations encode rich semantic and structural knowledge learned from internet-scale pre-training. However, the knowledge in these methods is implicitly defined and lacks explicit structure, making it difficult to interpret, control, or adapt to different manipulation tasks.

To address this limitation, an alternative approach is to explicitly define and integrate structured knowledge into the learning process. Object pose is one of the most fundamental types of knowledge~\cite{duan2017oneshotimitationlearning, tremblay2018deepobjectposeestimation, tyree20226dofposeestimationhousehold, vitiello2023one}. By incorporating pose information into imitation learning, methods like~\cite{track2act, spot, vitiello2023one} enable the policy to generalize across different object poses via demonstration replay. Nevertheless, such simplistic knowledge is unable to carry richer information, limiting its ability to address more complex manipulation tasks and generalize effectively to novel objects.

Some researchers have opted to use semantic keypoints to represent objects~\cite{di2024keypoint, fang2024keypoint, levy2024p3poprescriptivepointpriors, papagiannis2024r} to capture richer, more detailed information. These semantic keypoints are then utilized as input for imitation policies, improving task performance and generalization across various scenarios. However, despite their effectiveness in encoding object-centric knowledge, detecting keypoints can be challenging and is often influenced by factors such as self-occlusion, which can compromise the consistency of observations and, in turn, affect the policy's performance. 

To this end, we propose utilizing a keypoint graph to represent the object, ensuring more robust keypoint matching and enhancing knowledge-driven policy learning by providing a stable, structured representation that is less susceptible to visual disturbances such as occlusions. The most similar works to ours are~\cite{iga, vosylius2024instant}, which also abstract visual observations into a keypoint graph and perform in-context learning~\cite{brown2020language} based on the constructed graph. However, their keypoints are directly sampled from the point cloud using geometry-based sampling methods, overlooking semantic features in the image space, and potentially leading to inconsistent keypoint selections across frames.

\section{Method}

\subsection{Overview}
In this paper, we propose a new approach that can achieve knowledge-driven imitation learning, which leverages the semantic features and structural information of objects to achieve robust, generalizable, and sample-efficient robot learning. Our method begins by extracting a knowledge template of the target object from a single image. This knowledge template is then used to describe different objects within the same category by matching it to the object. Finally, we can train the knowledge-driven imitation policy based on the obtained object semantic knowledge, enabling the policy to adapt to varying conditions. The overall pipeline is illustrated in Fig.~\ref{fig:pipeline}.


\subsection{Category-Level Knowledge Template}

\begin{figure}[h]
    \centering
    \includegraphics[width=0.7\linewidth]{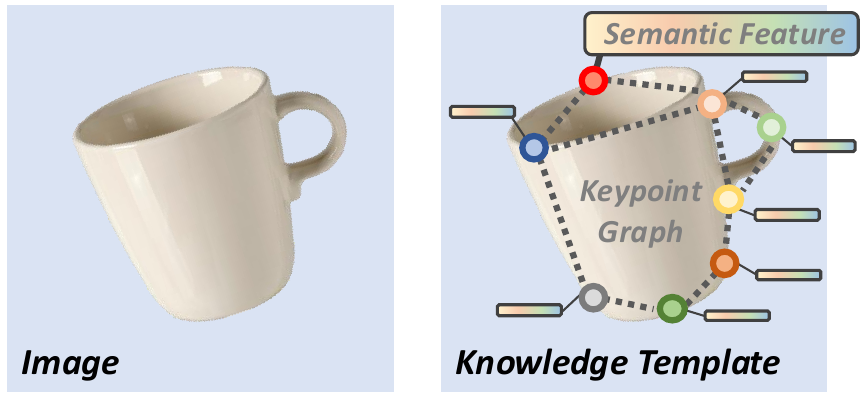}
    \caption{\textbf{Keypoint Graph as Knowledge Template.} The knowledge template is represented as a graph, explicitly encoding structural information, while each keypoint is associated with a semantic feature, implicitly conveying semantic information.}
    \label{fig:knowledge-template}\vspace{-0.4cm}
\end{figure}

Given a category $\mathcal{C}$, and an object $\mathcal{O}$ associated with the category, the system takes a single observation of the object to define a category-level knowledge template $\mathcal{T}^\mathcal{C}$ to abstract the common structural and semantic information shared by the objects within the category $\mathcal{C}$. Specifically, the template contains $K$ semantic keypoints,
$$
    \mathcal T^\mathcal{C} = \{ (\hat f_k, \hat p_k )\}_{k=1}^K,
$$ 
where $f_k \in \mathbb R^N$ denotes the semantic feature, $N$ denotes the feature dimension, $p_k \in \mathbb R^3$ represents the position of the $k$-th keypoint. In pratice, we treat the set of keypoints as a graph and consider the spatial relationships between keypoints during the matching process to ensure more effective template matching, as shown in Fig.~\ref{fig:knowledge-template}. The detailed information about the template matching will be introduced in Sec.~\ref{sec:matching}. We now describe how to obtain the semantic feature and how to sample the semantic keypoints as follows.

\paragraph{Semantic Feature Generation} To obtain the semantic feature $f$ of a specific point $p$, we first employ a pretrained semantic feature extractor. Given an RGB image $\mathcal{I} \in \mathbb R^{H\times W\times 3}$ of a certain object $\mathcal{O}$, a feature extractor $\mathcal F$ (e.g., DINOv2~\cite{oquab2023dinov2}) generates corresponding semantic features $\mathcal F(\mathcal{I}) = \{f\} \in \mathbb R^{H\times W\times N}$ for each pixel of the image. 
Combined with the depth image $\mathcal{D} \in \mathbb R^{H\times W}$ and camera parameters, we project pixels with their features into a 3D point cloud $\mathcal P = \{(f, p, c)\}$, where $p$ is the spatial position and $c\in \mathbb R^3$ is the color. 

\paragraph{Sematic Keypoint Sampling}
Then we need to select a subset of the point cloud $\mathcal P$ as the keypoints to form the knowledge template $\mathcal{T}^\mathcal{C}$. There are various methods to generate keypoints, such as sampling based on a specific strategy or annotating them manually by humans. These keypoints should have stable and distinctive features, and they must be evenly distributed across different parts of the object. Given the mask of the target object generated by SAM~\cite{kirillov2023segany}, one automated approach for keypoint selection is farthest point sampling. Directly using the position $p$ for farthest point sampling may fail to capture small but critical points, such as the rim of a mug or the handle of a drawer. Therefore, to ensure that the selected keypoints cover different parts of the object, we incorporate color $c$ and semantic feature $f$ to perform farthest point sampling. After $K$ keypoints are sampled, the semantic keypoints $\{ (f_{k}, p_{k})\}_{k=1}^K$ are obtained. We assume that the semantic features and structure of the keypoints can be generalized among objects in the same category, which means that this set of semantic keypoints can serve as the knowledge template $\mathcal{T}^\mathcal{C}$ of the category. 



\subsection{Knowledge Template Matching}\label{sec:matching}

After obtaining the knowledge template $\mathcal{T}^\mathcal{C}$ of category $\mathcal{C}$, we can register it to any new observation $(\mathcal{I},\mathcal{D})$ of object $\mathcal{O}^\prime$ to establish the relationship between the object in the new image and the template, where $\mathcal{O}^\prime$ can be unseen object in the category. We use the same semantic feature extractor to get semantic feature $f$ for each pixel, and projected each pixel to 3D point cloud $\mathcal P = \{(f, p,c)\}$. A simple way to find corresponding keypoint matching $m_k$ for $(\hat f_k,\hat p_k)\in \mathcal T^\mathcal{C}$ is the nearest point match based on feature distance~\cite{levy2024p3poprescriptivepointpriors, oquab2023dinov2,tang2023emergent}:
\begin{equation}
    m_k=\mathop{\text{argmin}}_{(f_i, p_i)\in \mathcal C} dis( f_i, \hat f_k),
\end{equation}
where $dis$ is the distance function, specifically the L2 distance.

Although the semantic features provided by the pretrained model have strong generalization ability, directly using the nearest point matching can still lead to limitations. One such scenario occurs when 
the features of the selected keypoints are similar, using the nearest feature matches may have a different order. And also when a keypoint is occluded during testing, the nearest feature matching will not work. Although the knowledge-based policy may learn to handle uncertainty, this uncertainty inherently increases the demand for training data and complicates model training.

To eliminate the ambiguity of keypoint matching, our proposed method incorporates keypoints to a template to provide structural information. And the method performs template matching instead of per point matching to achieve more robust matching. During template matching, apart from simply searching for the nearest keypoint, we also need to preserve structural consistency between detected and annotated keypoints. 
Formally, we aim to find an optimal template match $\mathcal M =\{m_k\}_{k=1}^K$ that
\begin{equation}
    \mathcal M = \mathop{\text{argmin}}_{(f_i, p_i)\in \mathcal C} \mathcal L_\text{feature} + \beta \cdot \mathcal L_\text{structure},
    \label{eq:target}
\end{equation}
where
$$
\left\{\begin{aligned}
    \mathcal L_\text{feature} &= \sum_{k=1}^K dis(f_i,\hat f_k),  \\
    \mathcal L_\text{structure} &= \min_{R,t,s} \sum_{k=1}^K  \|p_i- (s\cdot R\cdot \hat p_k + t)\|_2,
\end{aligned}\right.
$$
and $R,t,s$ denote rotation, translation, and scale, respectively; $\beta$ is the weight coefficient. 


To achieve this, we introduce a coarse-to-fine template matching pipeline that splits the optimizing problem into two parts.
We detail the template matching below.

\begin{figure*}
    \centering
    \includegraphics[width=0.95\linewidth]{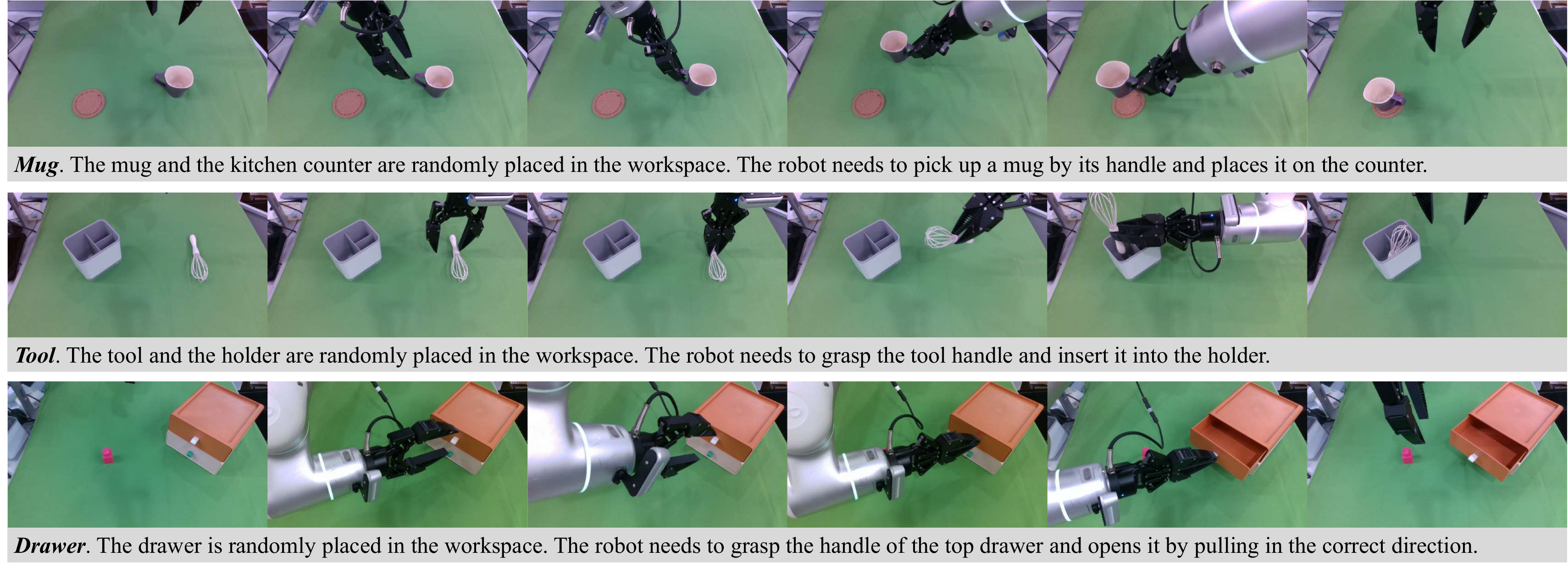}
    \caption{\textbf{Tasks.} We design three manipulation tasks (\textbf{\textit{Mug}}, \textbf{\textit{Tool}} and \textbf{\textit{Drawer}}) for real-world evaluations.}
    \label{fig:task}\vspace{-0.3cm}
\end{figure*}

\paragraph{Coarse Template Matching} 
Coarse template matching is responsible for efficiently performing an approximate estimation of the transformation between the knowledge template $\mathcal{T}^\mathcal{C}$ and the semantic point cloud $\mathcal{P}$.
In practice, it is a rigid point-set registration problem with scaling. If point correspondences are available, we can solve it using the Umeyama algorithm~\cite{88573}.

During the coarse matching process, the system needs to ensure that the matched keypoint have similar semantic features.
Specifically, we set a threshold $\delta_f$, considering pairs of points with feature distances below $\delta_f$ as correspondences. A more intuitive interpretation is that we redefine the feature distance as a penalty $\lambda$. Then, the objective function can be estimated as
\begin{align}
    \mathcal M = \mathop{\text{argmin}}_{(f_i, p_i)\in \mathcal C} \min_{R,t,s}
    \sum_{k=1}^K \|p_i- (s\cdot R\cdot \hat p_k + t)\|_2 + \lambda(f_i,\hat f_k),
\end{align}
where $\mathcal P$ is a penalty for mismatches
\begin{align}
    \lambda(f_i,\hat f_k) = \begin{cases}
        0 & dis(f_i,\hat f_k) \le \delta_f,\\
        +\infty& dis(f_i,\hat f_k) > \delta_f.\\
    \end{cases}
\end{align}

Considering the presence of occlusions, some keypoints may always be unmatched. To address this, we employ the RANSAC strategy for the matching process. This enables the algorithm to rely on a small number of accurately matched keypoints to infer the positions of occluded keypoints, enabling more robust template matching.

\paragraph{Fine Template Matching} 

While objects of the same class generally share a similar structure, there are still variations in the precise locations of keypoints. Coarse Template Matching provides an approximate transformation, but it oversimplifies feature distances, which is insufficient to capture the deformations. After obtaining this approximate transformation, further optimization of $\mathcal{L}_\text{feature}$ is required to capture the deformation.

To address this, after obtaining the approximate transformation, we further refine the keypoint positions. For each point, we search for better points within the neighborhood. Specifically, we solve the following optimization problem:
$$
    \mathop{\text{argmin}}_{(f_i, p_i)\in \mathcal C} dis(f_i,\hat f_k) + \beta \cdot \|p_i- (R \cdot \hat p_k + t)\|_2 + \lambda(p_i,\hat p_k),
$$
where the penalty $\lambda$ is defined as:
\begin{align}
    \lambda(p_i,\hat p_k) = \begin{cases}
        0 & dis(p_i,\hat p_k) \le \delta_p,\\
        +\infty& dis(p_i,\hat p_k) > \delta_p.\\
    \end{cases}
\end{align}
where $\delta_p$ is defined as the position distance threshold during fine matching.


Through the carse-to-fine template matching pipeline, we finally find an approximate matching for Eqn.~\eqref{eq:target}.









\subsection{Knowledge-Driven Policy}

To train a knowledge-driven policy via imitation learning, the proposed pipeline first generate knowledge templates for the relevant objects in the demonstrations, and perform knowledge template matching to the entire demonstration dataset. Once template correspondences $\mathcal M$ are obtained, the policy will be trained upon these, taking the matching results as inputs.
In practice, we treat the matching result of the knowledge template as a feature vector, which is then directly fed into a diffusion model to generate actions.
\begin{align}
    a = \pi_{\theta}(\mathcal{M},o_\text{other}),
\end{align}
where $\pi$ is Diffusion Policy and $o_\text{other}$ contains other necessary observations, including the robot proprioception information and gripper width. When the policy is deployed under novel conditions, like novel objects, the knowledge template will be matched to the novel object. And these matched keypoints can serve as anchors to deform objects in the current scene into template objects from the training set. At this stage, the policy trained on the template object perceives the novel object as a deformation of the template, allowing the actions learned from the template to be generalized to the new object. 

\section{Experiments}

\subsection{Setup}

\subsubsection{Platform} We use a Flexiv Rizon 4 robotic arm\footnote{https://www.flexiv.cn/product/rizon} with a Robotiq 2F-85 gripper\footnote{https://robotiq.com/products/adaptive-grippers} as our robot platform. The gripper fingers are replaced to TPU soft UMI finger~\cite{chi2024universal}. We use two third-person Intel RealSense D435i RGB-D cameras\footnote{https://www.intelrealsense.com/depth-camera-d435i/} for visual perception. One camera is positioned in front of the robot, facing the workspace, while the other camera, placed beside the robot, assists in capturing occluded areas. 

\subsubsection{Tasks} To evaluate the performance of our proposed knowledge-driven imitation learning method, we design three challenging real-world tasks that require precise localization, as shown in Fig.~\ref{fig:task}. The \textbf{\textit{Mug}} task demands accurate pose estimation, where even a 1 cm deviation can lead to failure; the \textbf{\textit{Tool}} task requires the policy to correctly determine the tool’s orientation and precisely grasp its center; and the \textbf{\textit{Drawer}} task involves learning complex and precise trajectories to successfully open the drawer.





\subsubsection{Data Collection} The human demonstrations are collected via teleoperation using a Force Dimension sigma 7 haptic device\footnote{https://www.forcedimension.com/products/sigma}. All data, including RGB-D images from two cameras, robot proprioception information, and gripper width, are recorded at a frequency of 10Hz. We collect 45 expert demonstrations for the \textbf{\textit{Mug}} task and 20 for the \textbf{\textit{Tool}} and \textbf{\textit{Drawer}} tasks, using a single object for each task to facilitate further generalization evaluations.

\subsubsection{Baselines} We select two baselines for evaluations in this paper: (a) \textbf{\textit{Diffusion Policy}}~\cite{chi2024diffusionpolicy}, a 2D image-based imitation learning policy that utilizes diffusion models to generate robot actions conditioned on image observation; (b) \textbf{\textit{P3-PO}}~\cite{levy2024p3poprescriptivepointpriors}, a novel policy in imitation learning that uses semantic correspondence to generate keypoints as observations. For fairness, we replaced the BAKU~\cite{haldar2024bakuefficienttransformermultitask} architecture used in P3-PO with the Diffusion Policy~\cite{chi2024diffusionpolicy} for policy learning. Depth information is directly obtained using the RealSense cameras. For each image captured by the camera, we manually select a similar number of keypoints. 
Although P3-PO employs the Co-tracker~\cite{karaev23cotracker} to track keypoints and achieve closed-loop control, we did not adopt this approach due to frequent keypoint tracking failure caused by the robot arm occluding the keypoints during large movements. Consistent with our method, we input the coordinates of the keypoints, the robot state, and the current task progress as observations into the Diffusion Policy, and the spatial augmentation is applied to both the observation and the actions.


\subsubsection{Evaluation Protocols} 

Each policy is evaluated across multiple consecutive trials per task. To ensure consistency, the initial states of the objects in the same trial are aligned in the view of the front camera. Success rates are used as the evaluation metric, with success determined based on the following task-specific criteria:

\begin{itemize}
    \item \textbf{\textit{Mug}}: The mug is grasped at the handle and finally stand upright and in contact with the coaster.
    \item \textbf{\textit{Tool}}: The tool is inserted into the holder with the handle facing downward.
    \item \textbf{\textit{Drawer}}: The drawer is opened by more than 3cm.
\end{itemize}

We allow the policy to execute until it reaches a long-term stationary state or encounters an irreversible issue, such as the mug tipping over.

\subsection{Implementations}

\subsubsection{Semantic Feature Generation} We use DINOv2 ViT-S/14~\cite{oquab2023dinov2} as our semantic feature encoder to generate semantic features. The raw semantic features from multiple cameras are projected into the robot's coordinate system using camera calibration results. To focus on task-relevant regions, we only consider points within a predefined workspace.

\subsubsection{Sematic Keypoint Sampling} For each object category, we establish a template containing 3-20 representative keypoints selected through either manual annotation or farthest point sampling. The selection of the specific number of points depends on the complexity of the object.

\subsubsection{Policy} The policy implementation follows the diffusion policy framework with a U-Net backbone~\cite{chi2024diffusionpolicy}. The input to the policy consists of the composite observation \( o = [o_{\text{template}}, o_{\text{robot}}, o_{\text{progress}}] \), where \( o_{\text{template}} \in \mathbb{R}^{3 \times K} \) represents the keypoint in the robot base coordinate, \( o_{\text{robot}} \) is the robot state, and \( o_{\text{progress}} \) indicates the task progress. To explicitly provide temporal awareness for the policy, we define current task progress as $o_{\text{progress}} = \text{current step} / \text{total steps}$. During inference, the total number of steps is the approximate average number of steps in the training set. Following~\cite{wang2024rise}, the robot state and action space are defined by the end-effector's 3D position, 6D rotation representation~\cite{zhou2020continuityrotationrepresentationsneural}, and gripper width. The policy predicts 20-step action trajectories. During training, we apply spatial augmentations with \( \pm 0.2 \, \text{m} \) translations and \( \pm 30^\circ \) rotations on both keypoint coordinates and end-effector poses to increase data diversity. The policy is trained for 2000 epochs with a batch size of 60, using 100 diffusion training steps and 20 inference steps for the DDIM scheduler~\cite{ddim}.


\subsubsection{Execution} During task execution, the gripper and robotic arm may occlude parts of the object, reducing template matching performance. To maintain consistent policy execution and improve speed, we use an ``open-loop'' control policy, where robot states are real-time while keypoint observations $o_\text{template}$ are obtained from the first frame. Despite being open-loop, experiments demonstrate that our method delivers promising results without the need for closed-loop correction. 

\subsection{Results}

\begin{table}[h]
    \centering
    \begin{tabular}{cccc}
    \toprule
    \multirow{2}{*}{\textbf{Method}} & \multicolumn{3}{c}{\textbf{Success Rate} $\uparrow$} \\
    \cmidrule(lr){2-4}
    & \multicolumn{1}{c}{\textbf{\textit{Mug}}} & \multicolumn{1}{c}{\textbf{\textit{Tool}}} & \multicolumn{1}{c}{\textit{\textbf{Drawer}}} \\
    \midrule
    Diffusion Policy~\cite{chi2024diffusionpolicy} & 3/20 & 5/25 & 9/20 \\
    P3-PO~\cite{levy2024p3poprescriptivepointpriors} & 14/20 &15/25 &11/20 \\
    \midrule
    Ours  &\textbf{16/20} &\textbf{22/25} &\textbf{15/20} \\
    \bottomrule
    \end{tabular}
    \caption{\textbf{Evaluation Results of Seen Objects.} Our method achieves the highest success rates compared to baselines across all tasks, demonstrating the effectiveness of leveraging the proposed knowledge template for imitation learning policies.}
    \label{tableseen}
\end{table}

Tab.~\ref{tableseen} reports the success rates of our method and baselines for previously seen objects. Compared to the Diffusion Policy, both our method and P3-PO show improved success rates across all tasks, demonstrating the effectiveness of incorporating knowledge. Despite the complexity of the tasks, our method achieves a high success rate with only a few expert demonstrations. While P3-PO offers an effective image encoding approach, its reliance on top-1 matching can lead to keypoint mismatches, especially when the object's pose changes. By selecting more keypoints or gathering additional expert demonstrations, the policy may learn to correct errors from noisy keypoints. In contrast, our template matching algorithm provides cleaner keypoints, resulting in superior performance in the policy trained with our method.

\begin{figure}[h]
    \centering
    \includegraphics[width=0.8\linewidth]{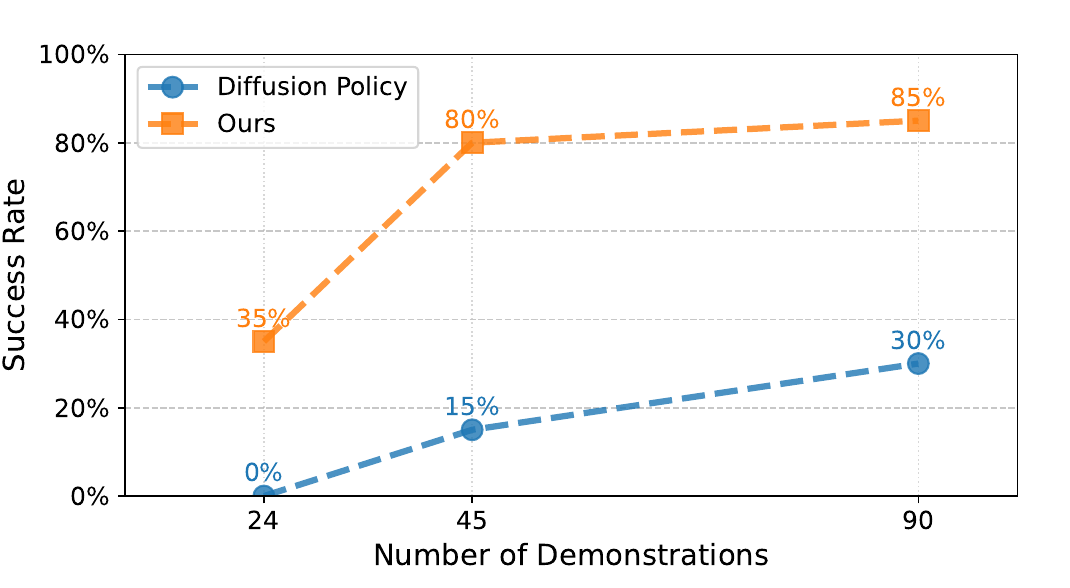}
    \caption{\textbf{Data Efficiency Comparisons.} Evaluation results of policies trained with varying numbers of demonstrations highlight the data efficiency of our proposed method compared to Diffusion Policy.}
    \label{dp}\vspace{-0.3cm}
\end{figure}

To investigate the data efficiency gained by incorporating knowledge, we collected an additional 45 expert demonstrations for the \textbf{\textit{Mug}} task. We then trained both Diffusion Policy and our method using 24, 45, and 90 demonstrations, respectively. Fig.~\ref{dp} shows the performance improvement of Diffusion Policy as the number of expert demonstrations increases. With approximately four times the number of expert demonstrations, the image-based Diffusion Policy begins to learn the object’s orientation and achieves a success rate comparable to that of our method with 24 demonstrations. This result also further confirms the critical role of prior knowledge when implementing imitation learning with limited expert demonstrations.

\subsection{Generalization Evaluations}

\begin{figure}[h]
    \centering
    \includegraphics[width=\linewidth]{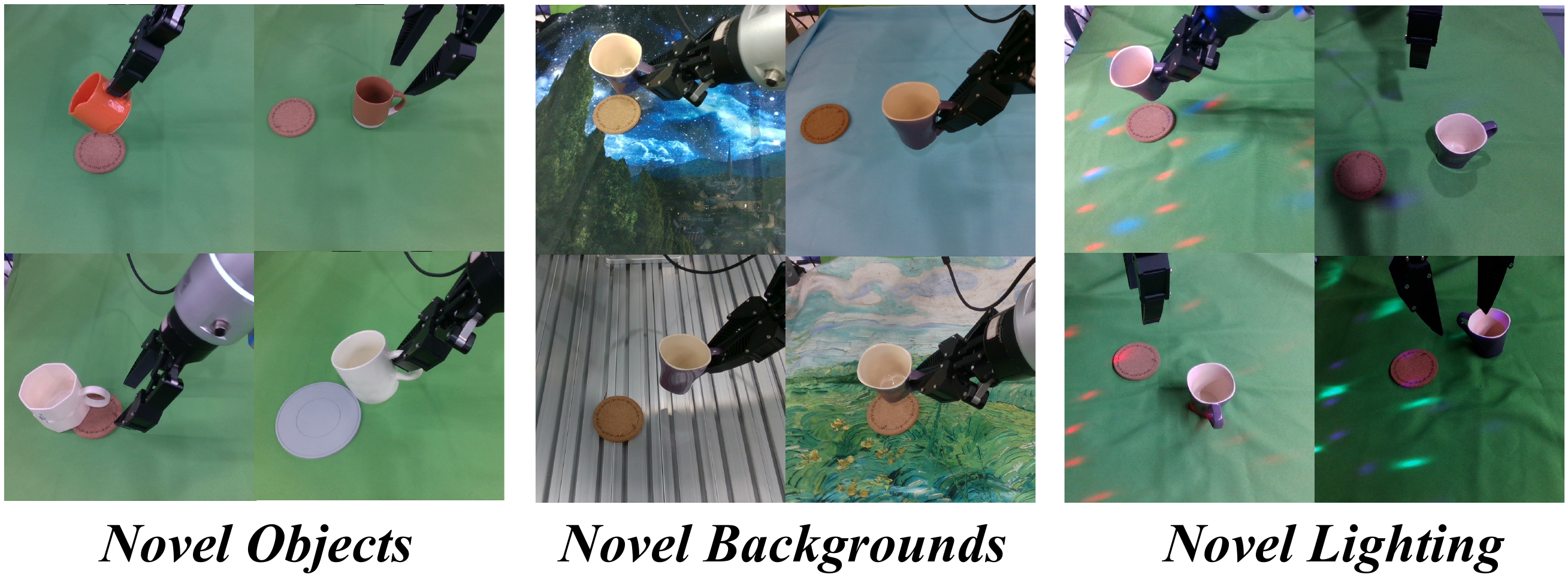}
    \caption{\textbf{Experimental Setup on Generalization Evaluations.} The policies are evaluated across three different variations: novel objects, backgrounds, and lighting conditions.}
    \label{fig:novel}
\end{figure}

To evaluate the generalization ability of our method across novel objects and environments, we introduce variations in both object types and environmental conditions during the evaluation, as shown in Fig.~\ref{fig:novel}. For the \textbf{\textit{Mug}} task, we test 5 unseen mugs along with 3 coasters. For the \textbf{\textit{Tool}} task, we evaluate 4 unseen tools, including a spoon, fork, and scraper. For the \textbf{\textit{Drawer}} task, we assess the method using an additional novel drawer. To further test generalization under environmental variations, we use the \textbf{\textit{Mug}} task as an example and apply four different backgrounds and several dynamic lighting conditions during the evaluations.

\begin{table}[htbp]
    \centering

    \begin{tabular}{cccc}
    \toprule    
    \multirow{2}{*}{\textbf{Method}} & \multicolumn{3}{c}{\textbf{Success Rate} $\uparrow$} \\
    \cmidrule(lr){2-4} 
    & \multicolumn{1}{c}{\textbf{\textit{Mug}}} & \multicolumn{1}{c}{\textbf{\textit{Tool}}} & \multicolumn{1}{c}{\textit{\textbf{Drawer}}} \\
    \midrule
    P3-PO~\cite{levy2024p3poprescriptivepointpriors} & \textbf{7/10} & 2/10 & \textbf{5/10}\\
    \midrule
    Ours & 6/10 & \textbf{7/10} & \textbf{5/10} \\
    \bottomrule
    \end{tabular}
    \caption{\textbf{Generalization Evaluation Results on Unseen Objects}. Our method demonstrates consistent and strong generalization performance on unseen objects across all tasks.}\label{tableunseen}\vspace{-0.3cm}
\end{table}

Tab.~\ref{tableunseen} shows the evaluation results for unseen objects. For the \textbf{\textit{Mug}} task, since 45 expert demonstrations have covered most poses of the mug, both P3-PO and our method generalized well to unseen objects. The inferior generalization ability of P3-PO on the \textbf{\textit{Tool}} task compared to our method is likely due to its reliance on the top-1 matching approach. When applied to tools, keypoints on the handle share similar shapes at various positions. Top-1 matching struggles to provide keypoints that are consistent with the location distribution and order of those in the training set. This results in poor performance during action generation. Additionally, the DIFT feature may fail to provide correct correspondences under different objects and poses.  In contrast, our method leverages semantic features to effectively locate the front and back of the handle, and the keypoint matching based on structured information allows for a more accurate estimation of the object's pose. Moreover, the structured information ensures that the location distribution and order of matched keypoints align closely with the training set, which helps improve the robustness of the action generation.

Tab.~\ref{environment} presents the success rate for novel background and lighting conditions for the \textbf{\textit{Mug}}  task, demonstrating that our method maintains a high success rate. This strong generalization ability can be attributed to our approach being built on a pre-trained semantic feature encoder, which enables adaptability to diverse environments.

\begin{table}[htbp]
    \centering
    \begin{tabular}{cccc}
    \toprule
    \multirow{2}{*}{\textbf{Method}} & \multicolumn{3}{c}{\textbf{Environmental Variations}} \\ \cmidrule(lr){2-4}
    & Original & Novel Background & Novel Lightning \\
    \midrule
    Ours & 16/20& 17/20& 10/20\\
    \bottomrule
    \end{tabular}
        \caption{\textbf{Generalization Evaluation Results on Environmental Variations.} Our method maintains strong success rates despite various environmental disturbances.}\label{environment}\vspace{-0.3cm}
\end{table}




\subsection{Ablations}

In this subsection, we investigate the contribution of each component in our method.

\subsubsection{Different Feature Extractors}

\begin{figure}[!h]
\centering
\captionsetup[subfigure]{justification=centering}

\begin{subfigure}[b]{0.48\columnwidth}
    \includegraphics[width=\linewidth]{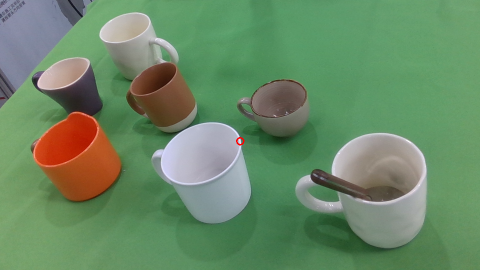}
    \caption{Source Image}
    \label{fig:sub1}
\end{subfigure}
\hfill
\begin{subfigure}[b]{0.48\columnwidth}
    \includegraphics[width=\linewidth]{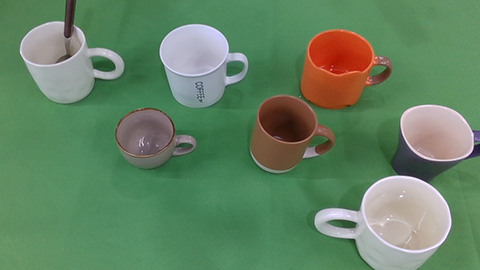}
    \caption{Target Image}
    \label{fig:sub2}
\end{subfigure}

\vspace{0.3cm} 

\begin{subfigure}[b]{0.48\columnwidth}
    \includegraphics[width=\linewidth]{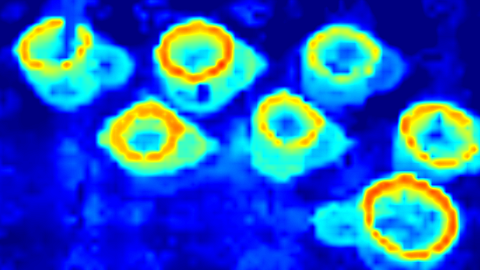}
    \caption{DINOv2~\cite{oquab2023dinov2}}
    \label{fig:sub3}
\end{subfigure}
\hfill
\begin{subfigure}[b]{0.48\columnwidth}
    \includegraphics[width=\linewidth]{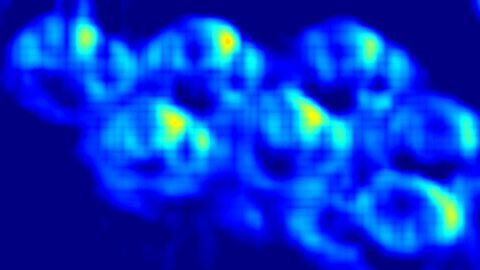}
    \caption{DIFT~\cite{tang2023emergent}}
    \label{fig:sub4}
\end{subfigure}

\caption{\textbf{Visualization of Correspondence by DINOv2 and DIFT.} The red circle indicates the selected point in the source image, while the heatmaps visualize similarity relative to this reference point.}
\label{fig:vis}\vspace{-0.3cm}
\end{figure}

The pretrained semantic feature extractor is a critical part of knowledge-driven imitation learning. Different feature extractors may provide different correspondence for the same selected point. Fig.~\ref{fig:vis} illustrates the point correspondence results across different semantic image encoders. The red circle indicates the selected point in the source image, while the heatmaps visualize similarity relative to this reference point.

As shown in Fig.~\ref{fig:vis}, DIFT~\cite{tang2023emergent} demonstrates a stronger emphasis on local shape features compared to DINOv2~\cite{oquab2023dinov2}, assigning discriminative feature values to regions with distinct orientations (e.g., the mug rim). In contrast, DINOv2 exhibits more homogeneous feature distributions. This homogeneity implies that DINOv2 could be preferable for template matching to improve correspondence accuracy.







\subsubsection{Template and Keypoint Matching}

To evaluate the effectiveness of template and keypoint matching, we rendered various objects from the Google Scanned Objects \cite{downs2022googlescannedobjectshighquality} in different poses via Blender. We then compare the average L2 error between the keypoints matched using different matching methods. As shown in Tab.~\ref{sim}, our matching method achieves a lower  average error compared to top-1 matching. To minimize potential interference from misalignment between the extracted features and the depth map, we manually removed the point cloud corresponding to the desktop in this experiment.

To further investigate the impact of template matching on policy learning, we assessed whether the matching process preserves the structural integrity of the keypoints. Specifically, we define a keypoint as ``matched'' if its matching error is less than 3 cm. Table~\ref{sim} demonstrates that our template matching algorithm significantly improves the matching rate, ensuring that the keypoint order remains consistent with the template.



\begin{table}[htbp]
    \centering
    \setlength\tabcolsep{3pt}
\begin{tabular}{ccc}
    \toprule
     \textbf{Matching Variants} &  \textbf{Average Error} (cm) $\downarrow$ & \textbf{Matched Rate} $\uparrow$ \\ \midrule
     Top-1 Keypoint Matching & 6.54 & 11.2\% \\
     Template Matching (\textit{ours}) &  \textbf{3.66} &  \textbf{59.2\%}\\
    \bottomrule
\end{tabular}
\caption{\textbf{Ablation Results of Template and Keypoint Matching.} Our template matching achieves a lower average error and a higher match rate compared to top-1 keypoint matching.}
\label{sim}
\end{table}



\subsubsection{Fine Template Matching}
\begin{figure}[h]
    \centering
    \includegraphics[width=0.7\linewidth]{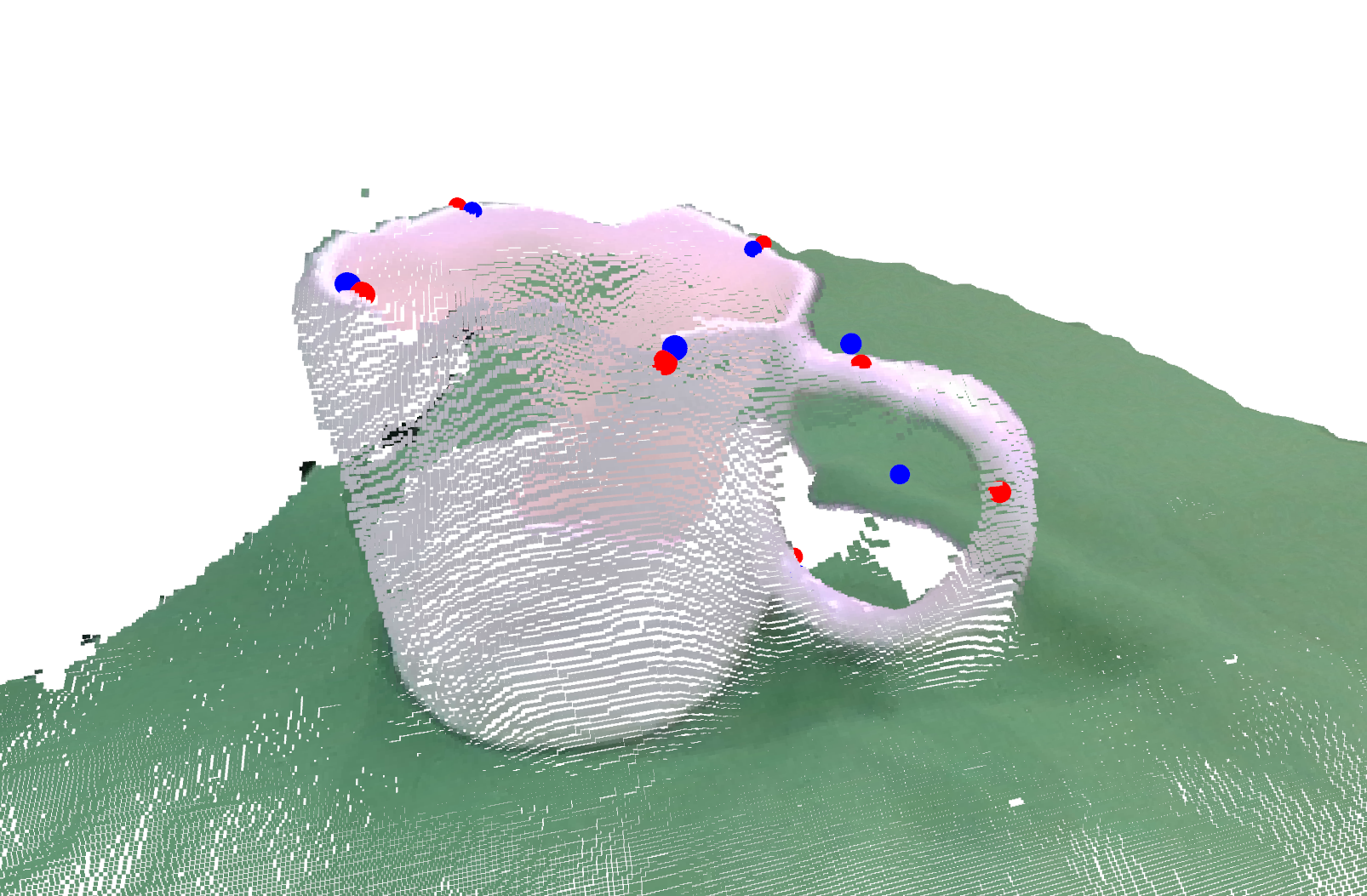}
    \caption{\textbf{Illustration of Fine Template Matching}. Fine template matching allows for the successful matching of mugs with handles of different sizes.}
    \label{fig:deformation}\vspace{-0.3cm}
\end{figure}

It is designed for unseen objects with different shapes. Fig.~\ref{fig:deformation} shows the result of matching a mug with a small handle to a mug with a large handle, where blue points are obtained by our coarse template matching and the red points are the deformation results. While coarse template matching provides the 6d-pose estimation of the unseen object, fine template matching enables more precise localization of grasp positions through  feature alignment. 

\section{Conclusions}

In this work, we introduce the knowledge-driven imitation learning system, a more efficient and generalizable robot manipulation learning method. We define a novel knowledge template format with a coarse-to-fine template matching algorithm, considering both the structural and semantic information.
We evaluate our knowledge-driven imitation learning method on three tasks. Experiment shows that the introduction of knowledge leads to a $54\%$ increase in the average success rate of Diffusion Policy. Our method outperforms Diffusion Policy with around 1/4 of the number of demonstrations. Our method is also robust to novel objects, novel backgrounds, and novel light conditions.


\printbibliography

\end{document}